\documentclass[10pt, conference, compsocconf]{IEEEtran}
\ifCLASSINFOpdf
\else
\fi
\usepackage{graphicx}
\usepackage{subfig}
\usepackage{balance}
\usepackage{hyperref}

\begin{document}
%
% paper title
% can use linebreaks \\ within to get better formatting as desired
\title{SGNet: A Super-class Guided Network for Image Classification and Object Detection}

% author names and affiliations
% use a multiple column layout for up to two different
% affiliations

\author{\IEEEauthorblockN{Kaidong Li$^{1}$, Nina Y. Wang$^{2}$, Yiju Yang$^1$, Guanghui Wang$^{3*}$}
\IEEEauthorblockA{$^1$ \textit{Department of Electrical Engineering and Computer Science, University of Kansas, Lawrence KS, USA, 66045}\\
$^2$ \textit{Department of Computer Science, University of Toronto, Toronto ON, Canada, M5S 1A1}\\
$^3$ \textit{Department of Computer Science, Ryerson University, Toronto ON, Canada, M5B 2K3}\\
$^*$ Corresponding author: wangcs@ryerson.ca}
}

% conference papers do not typically use \thanks and this command
% is locked out in conference mode. If really needed, such as for
% the acknowledgment of grants, issue a \IEEEoverridecommandlockouts
% after \documentclass

% for over three affiliations, or if they all won't fit within the width
% of the page, use this alternative format:
% 
%\author{\IEEEauthorblockN{Michael Shell\IEEEauthorrefmark{1},
%Homer Simpson\IEEEauthorrefmark{2},
%James Kirk\IEEEauthorrefmark{3}, 
%Montgomery Scott\IEEEauthorrefmark{3} and
%Eldon Tyrell\IEEEauthorrefmark{4}}
%\IEEEauthorblockA{\IEEEauthorrefmark{1}School of Electrical and Computer Engineering\\
%Georgia Institute of Technology,
%Atlanta, Georgia 30332--0250\\ Email: see http://www.michaelshell.org/contact.html}
%\IEEEauthorblockA{\IEEEauthorrefmark{2}Twentieth Century Fox, Springfield, USA\\
%Email: homer@thesimpsons.com}
%\IEEEauthorblockA{\IEEEauthorrefmark{3}Starfleet Academy, San Francisco, California 96678-2391\\
%Telephone: (800) 555--1212, Fax: (888) 555--1212}
%\IEEEauthorblockA{\IEEEauthorrefmark{4}Tyrell Inc., 123 Replicant Street, Los Angeles, California 90210--4321}}

% use for special paper notices
%\IEEEspecialpapernotice{(Invited Paper)}

% make the title area
\maketitle

\begin{abstract}
Most classification models treat different object classes in parallel and the misclassifications between any two classes are treated equally. In contrast, human beings can exploit high-level information in making a prediction of an unknown object. Inspired by this observation, the paper proposes a super-class guided network (SGNet) to integrate the high-level semantic information into the network so as to increase its performance in inference. SGNet takes two-level class annotations that contain both super-class and finer class labels. The super-classes are higher-level semantic categories that consist of a certain amount of finer classes. A super-class branch (SCB), trained on super-class labels, is introduced to guide finer class prediction. At the inference time, we adopt two different strategies: Two-step inference (TSI) and direct inference (DI). TSI first predicts the super-class and then makes predictions of the corresponding finer class. On the other hand, DI directly generates predictions from the finer class branch (FCB). Extensive experiments have been performed on CIFAR-100 and MS COCO datasets. The experimental results validate the proposed approach and demonstrate its superior performance on image classification and object detection.

\end{abstract}


% For peer review papers, you can put extra information on the cover
% page as needed:
% \ifCLASSOPTIONpeerreview
% \begin{center} \bfseries EDICS Category: 3-BBND \end{center}
% \fi
%
% For peerreview papers, this IEEEtran command inserts a page break and
% creates the second title. It will be ignored for other modes.
\IEEEpeerreviewmaketitle

\section{Introduction}
Recent years have witnessed the fast development of convolutional neural networks (CNNs) based models in computer vision tasks. Starting from 2012 \cite{krizhevsky2012imagenet}, CNN-based classifiers and detectors have quickly surpassed traditional models \cite{cen2019}\cite{wang2021salient}\cite{li20202}. In just a few years, the classifiers have already emerged to surpass human accuracy in several benchmark datasets \cite{he2015delving}\cite{zhang2021learning}. The success of CNN models is mainly owing to their abilities in extracting high-level semantic features from labeled data \cite{ma2020mdfn}\cite{cen2020deep}. CNN models have also been successfully applied in many other areas, like object detection \cite{ma2020location}, depth estimation \cite{he2018learning}, crowd counting \cite{sajid2020zoomcount}, and image translation \cite{xu2019toward}. 

What CNN-based models can achieve with simple class annotations is promising, but there are still weaknesses in this method. For the classification task, most models only take class names as input. All classes are treated equally without any correlation information. During training, the misclassification between ``chair'' and ``couch'' and that between ``TV'' and ``rabbit'' receive the same penalty. While in reality, ``TV'' and ``rabbit'' share almost no similarity, while ``chair'' and ``couch'' both belong to the furniture. This means the CNN models have to identify feature differences from multiple semantic hierarchies at the same time.
\begin{figure}[htp]
    \subfloat[Finer class labeling only]{
    \begin{centering}
        \includegraphics[width=0.4\columnwidth]{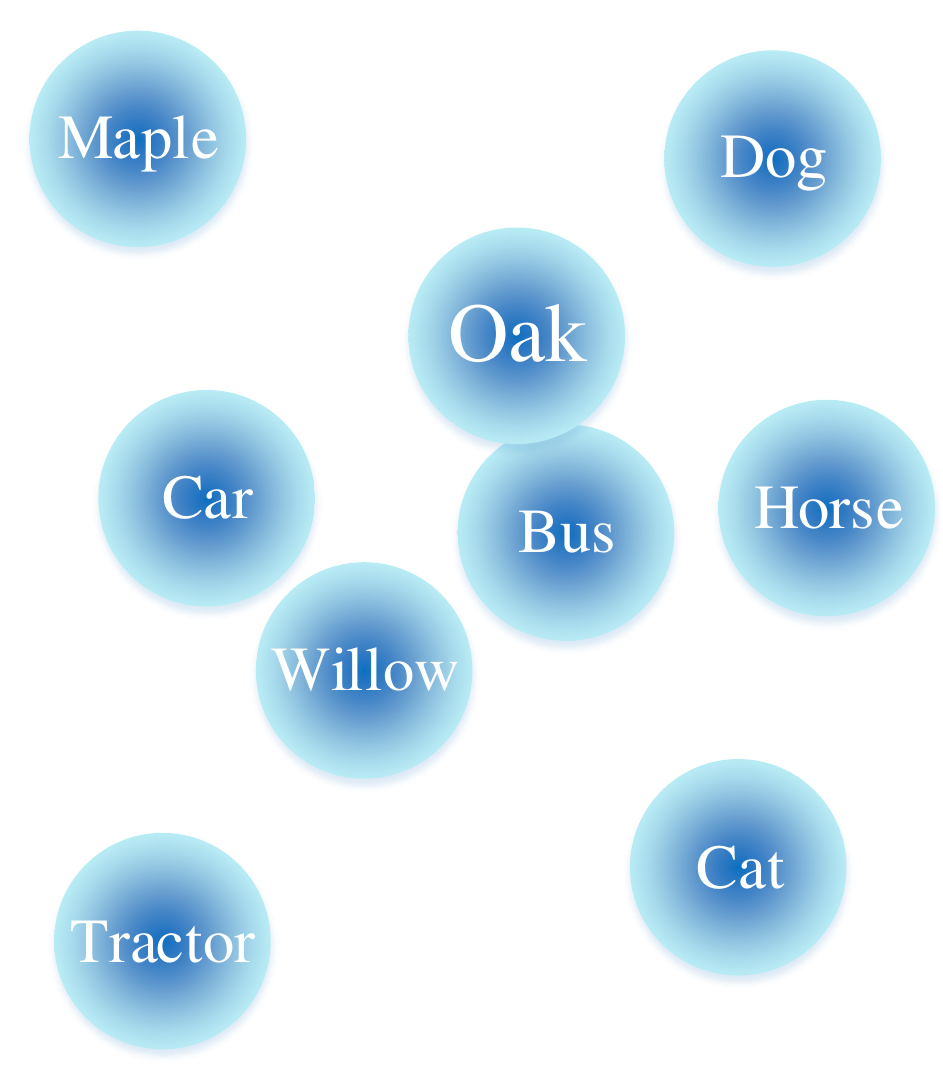}
        \label{plots:normal_cluster}
    \end{centering}}
    \subfloat[With super-class labeling]{
    \begin{centering}
        \includegraphics[width=0.4\columnwidth]{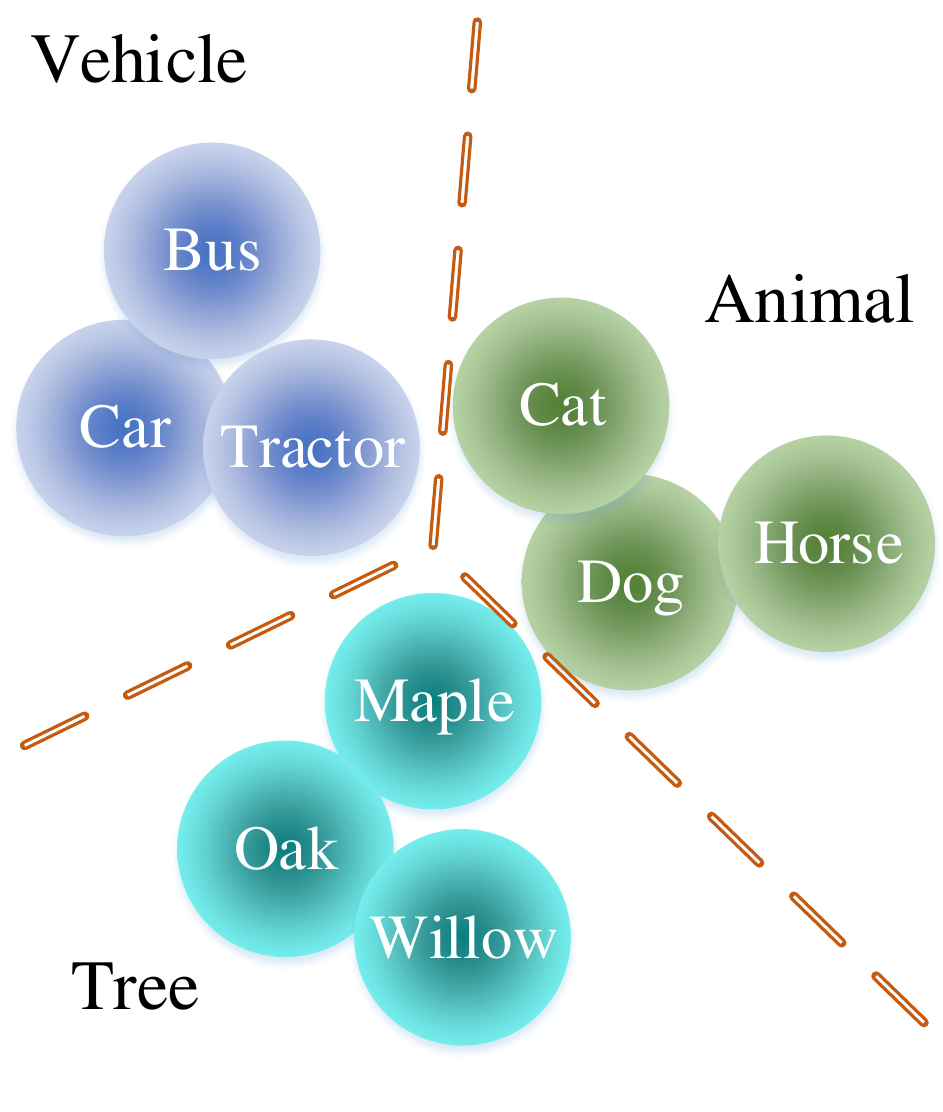} 
        \label{plots:super_cluster}
    \end{centering}}
	\caption{(a) The CNN models have to implicitly extract the semantic clustering of all classes. The distances among different clusterings depend on their semantic relations. (b) By introducing super-class labels, the model can first classify the features into high-level super-classes and then further classify them into finer classes using fine-grained details.} %By creating a 2-level hierarchy labels, (b) integrates human knowledge into the CNN models, making classification easier. }
	\label{plots:clustering}
\end{figure}

Our method is inspired by how human beings learn to identify an object. When creating categories, humans establish non-parallel, semantic relations between each class. Some classes, like ``animals'', are high on the semantic hierarchy and consist of high-level common features. Other classes might have very detailed features but are under a high-level super-class. As human beings, we tend to assign multiple labels to different levels of the semantic hierarchy. For example, when we see steak, we instinctively classify it as ``meat'', and upon closer inspection, the terms ``beef'' or ``ribeye'' might be assigned to the steak. Even if we are unsure what an object is, we are still able to classify it according to common features from a higher hierarchy. For example, even we have not seen a dog breed before, we're still very confident that it is a dog. This example shows how humans can generalize high-level super-class information to robustly recognize sub-class objects. %we may not know what a certain living organism is called, but they can most likely still be confident on whether the said organism is a "plant" or an "animal". 
%This example shows how the method by which humans classify objects is robust and able to handle many different circumstances.

In this paper, we propose a classification method that is similar to how humans identify an object. It first creates a set of super-classes based on original classes to establish a 2-level hierarchy. The model can simultaneously predict both super-classes and finer classes from separate branches. During training, a misclassified super-class will result in a loss for both branches, thus its value will be greater than a mere finer class error. The two branches share a majority of the backbone networks but still have their own individual portions. In this architecture, the super-class portion focuses on general features shared by super-class categories, while the finer class portion can handle detailed fine-grain attributes.

In this way, we are able to utilize the knowledge base humans have obtained to help the model learn the semantic clustering of classes. The current classification training only provides the network with parallel labels without explicit semantic information. As illustrated in Figure~\ref{plots:clustering}, the difficulty of extracting the semantic clustering can be alleviated by ``telling'' the model the 2-level hierarchy labels. Our method is also promising to integrate human semantic knowledge into deep learning models.
In this paper, we propose a super-class guided network (SGNet) and verify its effectiveness on both image classification and object detection tasks. Extensive experiments show that our proposed approach can consistently boost the performance of existing models with only a little overhead. The proposed method can be adapted to any existing image classification or object detection networks. %According to our experiments, we increase classification accuracy for CIFAR-100 \cite{krizhevsky2009learning} by $0.69\%$ and detection AP for MS COCO \cite{lin2014microsoft} from $28.6\%$ to $29.2\%$. At the same time, we still keep the model time efficient. Our SGNet still runs at 413 FPS for image classification and 23 FPS for object detection using VGG-16 as backbone network. 

The main contribution of this paper include:

\begin{enumerate} %\setlength\itemsep{-2pt}
	\item Inspired by the human cognition system, we propose a super-class guided architecture that consists of a super-class branch (SCB) and a finer class branch (FCB). The features from SCB with high-level information are fed to FCB to guide finer class predictions.
	\item By introducing high-level information and grouping existing finer classes into super-classes, current labels can be easily modified to train both SCB and FCB for better performance in detection and classification.
	\item The proposed SGNet can be directly applied to most image classification and object detection models. Extensive experiments demonstrate its superior performance over existing models.
\end{enumerate}

The source code of the proposed network can be accessed at \url{https://github.com/rucv/SGNet}.

\section{Related Work}
%Our method is inspired by some other attempts to integrate additional human knowledge into training processes. The strength of our work is demonstrated by improving baseline performance while requiring minimal additional labels and computation resources. 

\subsection{Image Captioning and Attributes}
The goals of image captioning and detection by attributes can vary to a large extent. However, at certain stages, they both aim to train the models to interpret the images in a descriptive way. Image captioning models \cite{liu2021vocabulary, you2016image, anderson2018bottom} combine computer vision and natural language processing techniques. By providing captions during the training phase, we expect the model not only to learn what the objects are, but also to understand the details of and interactions between those objects. Yao \textit{et al.} \cite{yao2017boosting} proposed an architecture to boost caption performance by incorporating attributes. 

Attributes have long been used in the computer vision community. Back to around 2010, some works \cite{farhadi2009describing} were published to generate attribute detectors. Later on, a popular director, zero-shot learning predicts object instances whose classes might not be in the training dataset. These detectors \cite{al2016recovering,lampert2013attribute} often employ the attribute detector as the first stage and then predict the class categories by finding the category with the most similar attribute set. 

Both of these two areas try to integrate human knowledge into deep learning models to obtain robust results. However, they have a common drawback: both methods require a specific set of annotations. For each object, they need at least several words to form the caption or attributes,  which limits the amount of data accessible, unlike an object detection ground truth, which consists of only a bounding box and a class label. Moreover, captions and attributes are more subjective, which make the annotations error-prone, which further limits the applications.

\subsection{Hierarchical Class Labeling}
Read \textit{et al.} \cite{read2013multi} uses conditional dependency information from classifiers' error vector to generate optimal super-class partitioning. Then, a common multi-dimensional ensemble method is used to predict the final result. Zhou \textit{et al.} \cite{zhou2018deep} applied the super-class idea on a training dataset with unbalanced class distribution. They first partitioned the classes into super-classes to create a relatively balanced distributed dataset, which helps minority classes benefit from abundant samples under the same super-class. A weight matrix is applied to put attention on features important to a specific super-class for final prediction. 

These works generate super-classes using statistic distribution from one dataset, which makes them dataset dependant. In addition, to calculate the statistic distribution, complex algorithms will get involved to complete the training process.
Roy \textit{et al.} \cite{roy2020tree} created an incremental CNN learning model with a tree structure. This model can evolve after the base model has completed training. When a new class sample is provided, the model can assign the new class to available nodes or even create new branches for the new class depending on the output. Therefore this method can add new classes without abandoning the existing classification ability.

Other methods \cite{wanglearning, yan2015hd} directly use human knowledge to create the super-class clustering. Wang \textit{et al.} \cite{wanglearning} proposed a CNN architecture with separate branches for super-classes and finer classes with some shared layers. It predicts super-class results if the finer class branch is not confident and finer class results when the confidence score from the finer class is above a certain threshold. Yan \textit{et al.} \cite{yan2015hd} also created a similar architecture. But the model has an individual branch for each super-class and predicts super-class results first. Once the super-class result is out, the model only uses the corresponding branch to predict the finer class. 

Different from our methods, these models don't take advantage of the high-level features from super-classes in finer class prediction. In our proposed approach, the high-level features are concatenated to the finer branch features. Therefore the finer class branch can focus on extracting fine-grained features. Also in \cite{yan2015hd}, each super-class has its own CNN branch, which will result in larger model size.

\section{The Proposed Method}\label{sect:method}
We propose a CNN architecture that can be applied to both classification and detection tasks. By adding a super-class branch (SCB), it is designed in a way that can be directly plugged into any existing model. The new branch is trained together with the original classification to guide the finer class training while learning the super-class information. The overall loss is calculated by summing the losses from two branches and backpropagates at the same time.

During inference, we experiment two different setups
\begin{enumerate} %\setlength\itemsep{-3pt}
	\item \textbf{Two-step inference} (TSI) first predicts super-class, then finds the highest confidence score in the corresponding finer classes. 
	\item \textbf{Direct inference} (DI) directly generates classification results from the finer class branch.
\end{enumerate}

TSI takes advantage of the higher accuracy in super-class prediction. Therefore, even if the final finer class prediction is not correct, it tends to avoid serious mistakes. The DI setup skips the computation in SCB which optimizes the inference time. In addition, it achieves better performance compared to the baseline, because during training, the SCB guides the model to extract the relations among different classes. The details will be discussed in Section \ref{Sect:exp}.

The key contribution of our proposed methods is the addition of the SCB, which shares most layers with the original CNN architecture and uses the common cross entropy loss in training. The detailed architecture will be discussed next.
\begin{figure}[t]
	\includegraphics[width=\columnwidth]{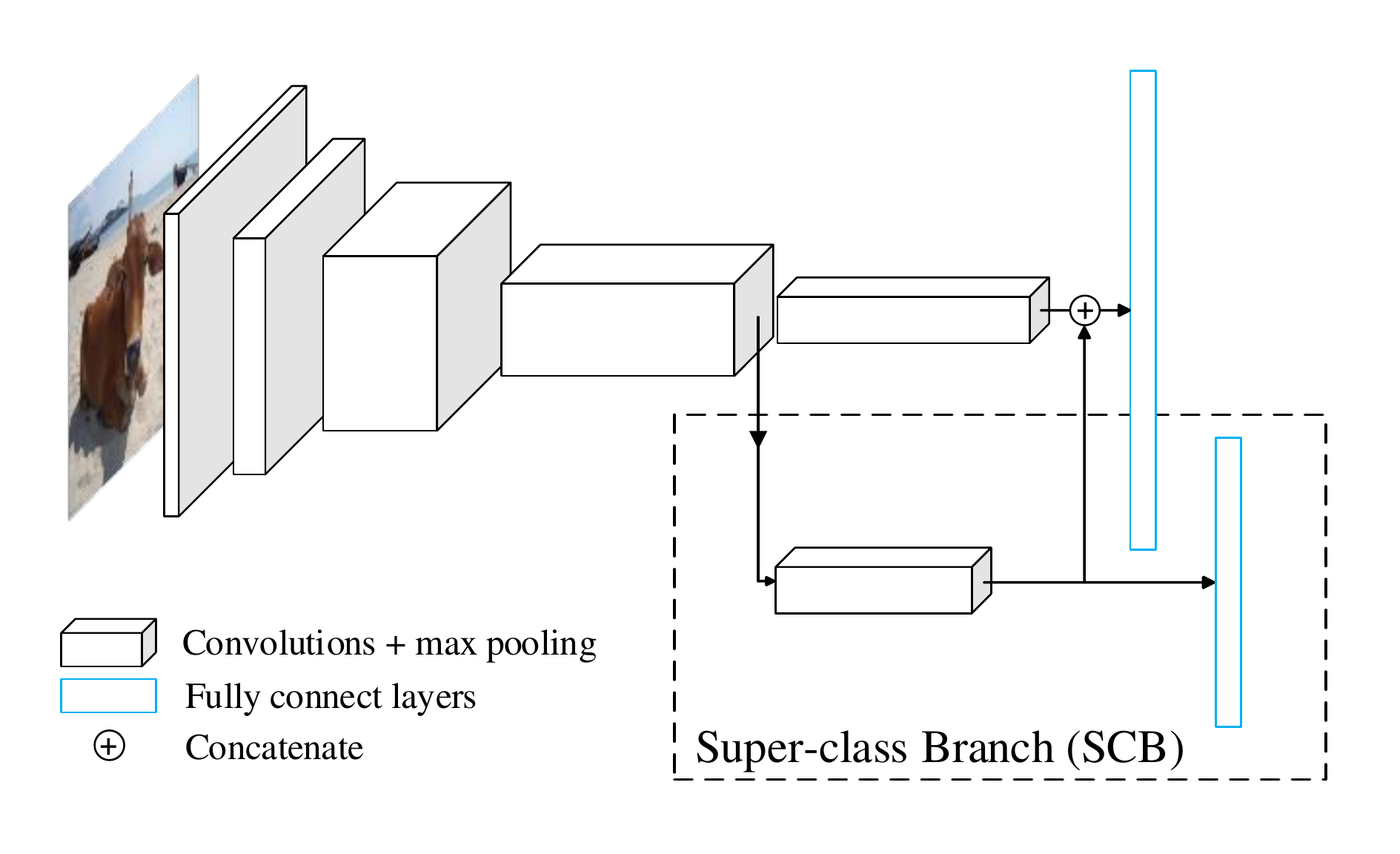}
	\centering
	\caption{\textbf{The architecture of SGNet.} SCB is in the dashed rectangle. It takes high-level features from the original networks as input and predicts super-class results. The features from SCB are then concatenated back to the original network to guide finer class prediction. SCB usually has fewer layers than its corresponding finer class branch.}
	\label{fig:our_arch_in1}
\end{figure}

\subsection{Super-class Branch for Classification}
A vast majority of recent CNN-based image classification models have a common architecture as shown in Figure~\ref{fig:our_arch_in1} without the SCB in the dashed box. They are a cascade of convolution layers to extract semantic information with downsampling layers inserted at certain levels, followed by several fully connected layers to predict image classes at the end. Most downsampling is done by max pooling, \textit{e.g.} VGG \cite{simonyan2014very} and ResNet \cite{he2016deep}. Some recent models achieve the same effect by applying convolutional layers with stride $>1$, like SqueezeNet \cite{iandola2016squeezenet} and MobileNets \cite{howard2017mobilenets}. Our method can be adapted to both of these architectures.

Our idea can be generalized to any classification model. The basic structure is shown in Figure~\ref{fig:our_arch_in1}. The added SCB takes one hidden layer of the main network as input. After going through a separate cascade of convolutional layers, downsampling layers, and fully connected layers, it predicts super-class confidence scores. The SCB's final feature maps, which is the input of SCB's fully connected layers, are concatenated back to the final feature maps in the original branch to generate finer class scores.

The input of SCB is taken after one of the last several downsampling layers in the main network, which contains high-level information since it is already towards the end of the architecture. The exact downsampling layer depends on the depth of the main network. For example, in VGG-16, we select the second last pooling layer to start the SCB. 

The number of convolutional layers in the SCB is designed to be shallower than the rest of the original, and there should be an equal number of downsampling layers. Detecting super-class is an easier task compared to finer classes. The feature differences among super-classes are much greater than those in finer classes. A shallower network can achieve faster performance, and it will be sufficient to guide the finer classification. The downsampling layers should match since the feature maps from both branches will be concatenated. Therefore, the feature maps have to share the same dimensions. 

The general architecture in the SCB is similar to a common CNN classification network. It consists of several convolutional layers followed by fully connected layers to make super-class prediction. The input of the fully connected layers, which is the final feature maps generated by SCB, will be concatenated to the feature maps of the finer class branch. The SCB feature maps contain information about common features within the same superclass. It can help the original branch only to focus on finer details to distinguish between categories within the same super-class.  

\subsection{Super-class Branch for Detection}
\begin{figure}[tp!]
    \subfloat[Two-stage detectors]{
    \begin{centering}
        \includegraphics[width=0.9\columnwidth]{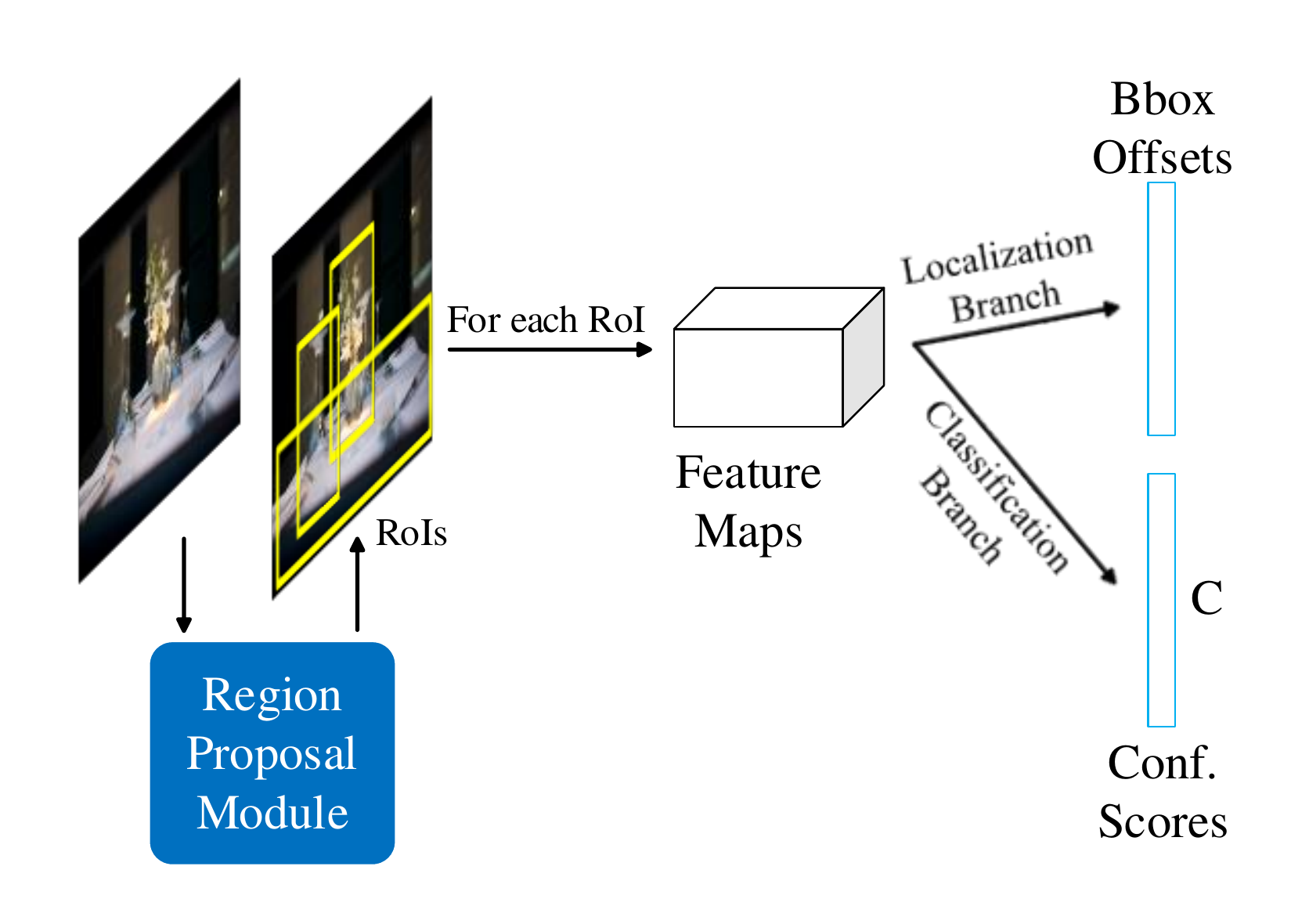}
        \label{plots:ts_det}
    \end{centering}}
    \hfil
    
    \subfloat[One-stage detectors]{
    \begin{centering}
        \includegraphics[width=0.9\columnwidth]{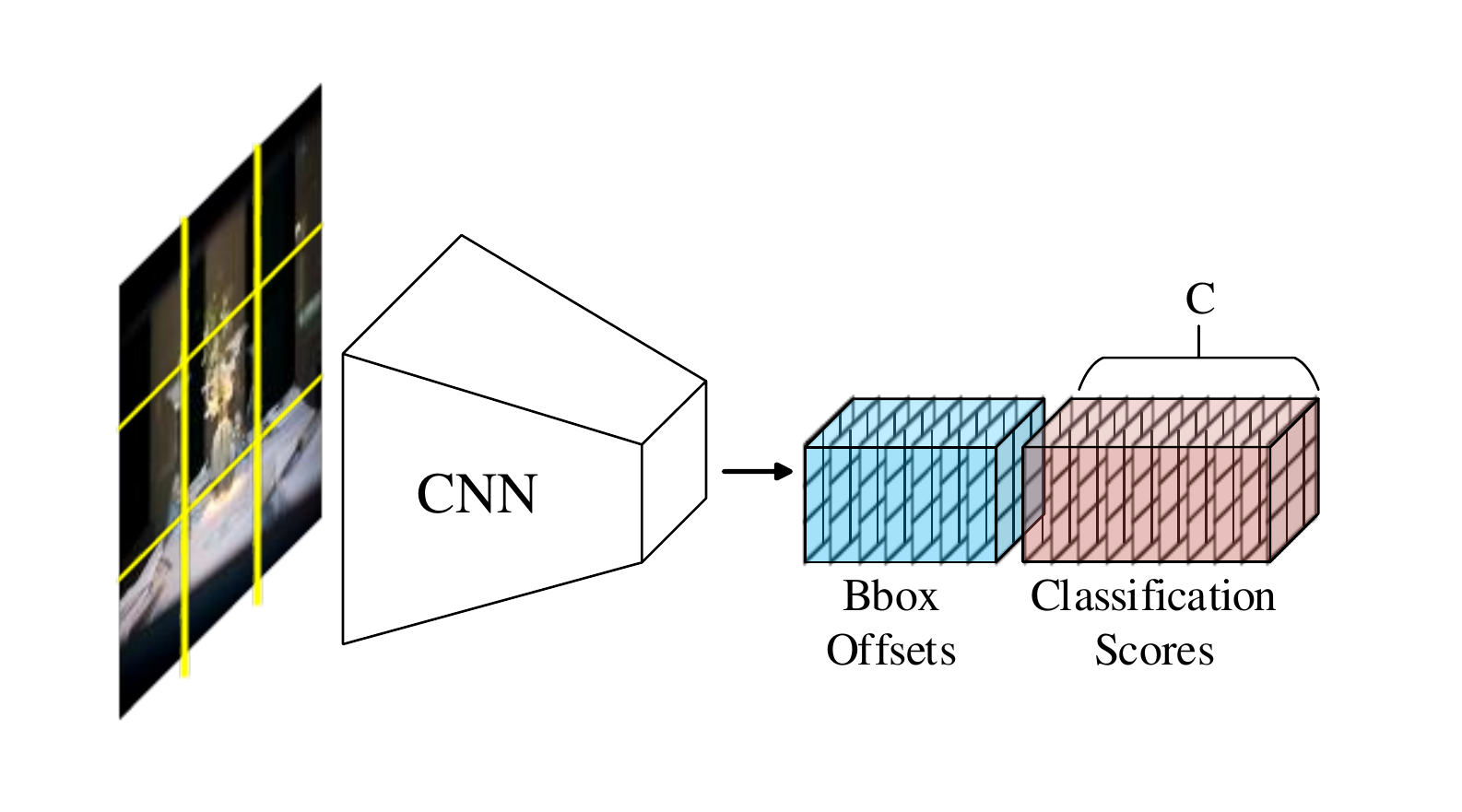} 
        \label{plots:os_det}
    \end{centering}}
    \hfil
    
	\caption{ \textbf{The architecture difference for different types of detectors} $C$ in both \protect\subref{plots:ts_det} and \protect\subref{plots:os_det} is the number of classes.}
	\label{plots:detectors}
\end{figure}

Generally speaking, there are two types of detectors, two-stage detectors and one-stage detectors. They have different architectures. Two-stage detectors, like the RCNN family \cite{girshick2015fast,pramanik2021granulated}, first propose region of interest (RoI). Then, they extract feature maps for the RoIs, from which bounding-box offsets and confident scores are calculated. Unlike two-stage detectors, one-stage detectors don't have a region proposal stage. Instead, most of them, like SSD-based detectors \cite{liu2016ssd, zhang2018single}, have predefined shapes (commonly referred to as anchors) evenly distributed in images using grid cells. The detectors directly generate offsets and confidence scores for the anchors. The architectures are shown in Figure~\ref{plots:detectors}.

\begin{figure}[tp!]
	\includegraphics[width=\columnwidth]{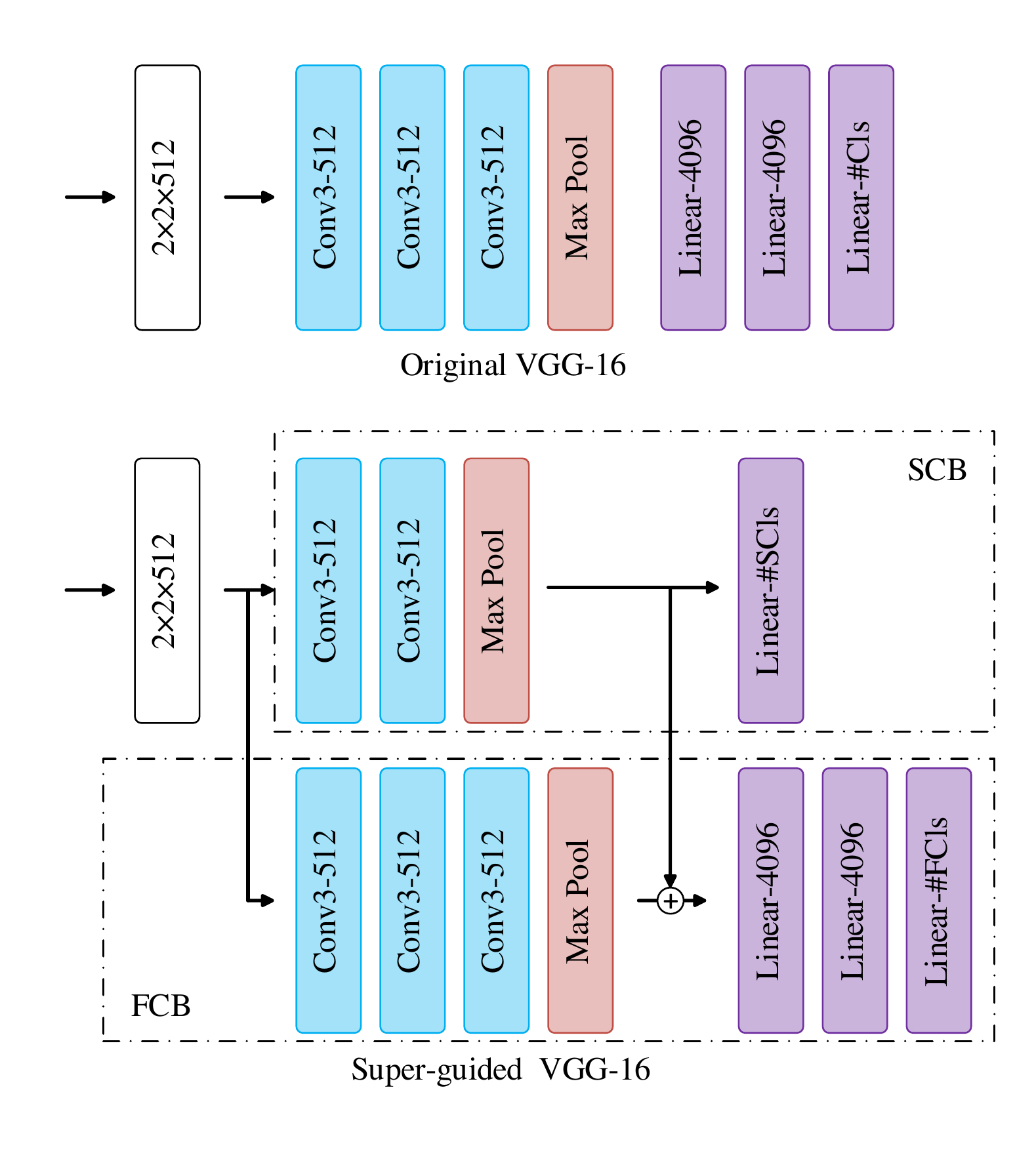}
	\centering
	\caption{\textbf{Partial architecture of super-class guided model in VGG-16.} The modification is after the 4th max pooling layer (10th layer) of the original VGG-16 network. The feature map at this stage is a $2\times2\times32$ tensor. \textbf{Upper}: The original model has 3 convolutional layers and one max pooling layer, followed by 3 fully connected layers. \textbf{Lower}: Only 2 convolutional layers and 1 fully connected layer are used in super-class branch (SCB) for the easier super-class prediction. The finer class branch (FCB) has the same depth as the original architecture. The input of FCB's fully connected layers is the concatenation of the feature maps from SCB and FCB.  }
	\label{fig:vgg_arch}
\end{figure}

Although we only focus on the classification branch, there are still lots of difference among the structures. For example, Faster RCNN \cite{ren2015faster} uses fully connected layers to produce confidence scores from a single feature layer, while SSD \cite{zhang2018single} applies convolutional layers to multi-layers from the backbone network. To generalize our approach, we designed a method that will work for all existing detectors. As shown in Figure~\ref{plots:detectors}, all models output $C$ scores for each area of interest (RoI for two-stage models and anchor for one-stage models) where $C$ is the number of finer classes $C_{FC}$. For super-class guided method, we simply make $C = C_{SC}+C_{FC}$. After the model produces the vector $V$ of length $C$, it divides the vector into two parts, $V_{SC} = V[0:C_{SC}]$ and $V_{FC} = V[C_{SC}:C_{SC}+C_{FC}]$,which are responsible for super-class and finer class predictions respectively.
\begin{table*}[t]
	    \centering\footnotesize
        \begin{tabular}{l|lllll}
            Super Class & \multicolumn{5}{c}{Finer Class} \\ \hline
            aquatic mammals & beaver & dolphin & otter & seal & whale  \\ 
            fish & aquarium fish& flatfish&ray& shark& trout \\
            flowers &orchids& poppies& roses& sunflowers& tulips\\
            food containers&	bottles& bowls& cans& cups& plates\\
            fruit and vegetables&	apples& mushrooms& oranges& pears& sweet peppers\\
            household electrical devices&	clock& computer keyboard& lamp& telephone& television\\
            household furniture&	bed& chair& couch& table& wardrobe\\
            insects	&bee& beetle& butterfly& caterpillar& cockroach\\
            large carnivores&	bear& leopard& lion& tiger& wolf\\
            large man-made outdoor things&	bridge& castle& house& road& skyscraper\\
            large natural outdoor scenes&	cloud& forest& mountain& plain& sea\\
            large omnivores and herbivores	&camel& cattle& chimpanzee& elephant& kangaroo\\
            medium-sized mammals&	fox& porcupine& possum& raccoon& skunk\\
            non-insect invertebrates&	crab& lobster& snail& spider& worm\\
            people&	baby& boy& girl& man& woman\\
            reptiles&	crocodile& dinosaur& lizard& snake& turtle\\
            small mammals&	hamster& mouse& rabbit& shrew& squirrel\\
            trees&	maple& oak& palm& pine& willow\\
            vehicles 1	&bicycle& bus& motorcycle& pickup truck& train\\
            vehicles 2&	lawn-mower& rocket& streetcar& tank& tractor\\
        \end{tabular}
         \vspace{3mm}
        \caption{\footnotesize Super-class and finer class groups for CIFAR-100}
        \label{tab:cifar_sc_fc}
	\vspace{-10pt}
\end{table*}

\subsection{Super-class Branch During Training and Inference}
%\noindent
\textbf{Training: } During training, the setup is straightforward. The classification ground truth for super-classes is converted from their original finer class annotations. The loss for super-class prediction and finer class prediction is calculated individually using cross-entropy loss. The new classification loss is defined by: 

\begin{equation}
\label{comb_loss}
Loss = (1-\alpha) Loss_{FC} + \alpha Loss_{SC}
\end{equation}

where $\alpha$ ($\in (0, 1)$) determines which loss to focus on. For example, a greater $\alpha$ trains super-class branches faster. The features generated from backbone networks are biased more towards super-classes.
\\~\\

%\noindent
\textbf{Inference: } At the inference stage, the basic architecture is the same as training. The SCB first generates the super-class confidence scores. Then, the original branch produces the finer class confidence scores. To obtain the final prediction, we experimented two setups. 

\begin{enumerate} 
	\item \textbf{Two-step inference} (TSI) predicts super-class by finding the maximum confident score from SCB. Then, it takes the corresponding finer class scores into consideration. We notice there are sometimes conflicts between the super-class and finer class predictions if we analyze the two branches individually. This setup uses the more accurate super-class results to guide finer class tasks. Thus, the finer class branch only needs to choose between categories within one super-class. The final confidence score is calculated by applying softmax layer to only the corresponding finer class outputs from fully connected layers.
	Using this setup, even when the final finer class predictions are incorrect, we still likely have the correct super-class results, thus resulting in less serious consequences. 
	
	\item \textbf{Direct inference} (DI) directly generates classification results from the finer class branch without even computing the super-class prediction. The features generated by SCB are concatenated to the original features to produce the final confidence scores. Any computation starting from fully connected layers in SCB is discarded. 
	This setup saves the time that is otherwise needed to go through the fully connected layers in SCB and the algorithm to find finer classes corresponded to the super-class prediction. And during training, the finer class branch learns across all finer classes, not limited to just the classes within one super-class. Therefore, the original finer class branch has already learned how to work with SCB features to yield the correct class prediction on its own. 
	
\end{enumerate}

\section{Experiments} \label{Sect:exp}
We conduct our experiments for two tasks, image classification and object detection. We use VGG-16 \cite{simonyan2014very} as the base network for both tasks.  It is chosen because it is a simple and widely-used network. For $\alpha$ value in Equation~\ref{comb_loss}, we choose $0.5$ so that SCB and FCB have equal contribution to the classification loss. For image classification, results are tested on CIFAR-100 \cite{krizhevsky2009learning}, which consists of $50k$ training images. Object detection task uses MS COCO \cite{lin2014microsoft}. The detailed information will be presented in the corresponding subsections.

\subsection{Classification Dataset}
\noindent
\textbf{CIFAR-100} is a dataset with tiny images of $32 \times 32$ resolution. It has a total of $60k$ images, with $50k$ in the training set. They are evenly distributed among $100$ finer classes, with $600$ images each. This dataset is chosen because it also provides its own 20 super classes, as shown in Table~\ref{tab:cifar_sc_fc}. Each super class has 5 finer classes. The evenly distributed pattern eliminates other factors and helps us to analyze our modified architecture. To evaluate the performance, we simply use the accuracy calculated from all test images.

The idea of super-class is to create a hierarchy that classes at each level should have the same semantic level. From Table~\ref{tab:cifar_sc_fc}, we can see the super-class categories in CIFAR-100 are not optimal. For example, \textit{trees} and \textit{small mammals} are obviously not at the same level. In fact, \textit{trees} is a level higher in ImageNet \cite{deng2009imagenet}. Further performance improvement is expected under well-designed hierarchy. But in this paper, we just implement the basic version to evaluate the effect.

\subsection{Classification Model Architecture}
In our image classification task, we choose VGG-16 \cite{simonyan2014very} as the starting point. The comparison between the original architecture and our super-class guided architecture is shown in Figure~\ref{fig:vgg_arch}. Most of the architecture is kept the same. Our modification starts after the 4th max pooling layer. The super-class branch is shallower than the original network, while the finer class branch (FCB) has the same depth as the original network. The difference in the FCB is the input of the fully connected layers. The concatenation results in a 1024-channel feature map, while the original network only has 512 channels. We design a 1-level fully connected layer in SCB since predicting super-class is comparatively easier. In the next subsection, we will show that this shallower network has satisfactory results. 

The learning rate is set to be $0.1$ initially for all training. We decay the learning rate at epoch $60, 120$, and $160$ with a rate of $0.2$. We also adopted a training strategy, \textit{warmup} \cite{he2016deep, goyal2017accurate}. By using a less aggressive learning rate at the start, it prevents the model from getting unstable at the beginning stage of training. In our experiments, we set the first epoch as the warmup stage, and used a batch size of $128$. 

\subsection{Classification Results}
We report the results from original VGG-16, SGNet with two-stage inference (TSI) and direct inference (DI) in Table~\ref{tab:cifar_accu}. All the experiments are conducted on an NVIDIA TITAN Xp GPU. The performance is compared using top-1 accuracy. Other metrics are also listed to thoroughly evaluate our proposed SGNet. 

\begin{table}[t]
	    \centering\footnotesize
        \begin{tabular}{l|llll}
            Model       & Accuracy ($\%$) & Epoch & Inference Time & \# Params \\
            \hline
            VGG-16      & 72.15           & 197   & 2.06ms            & 34.0M \\ 
            SG with TSI & 72.78           & 195   & 2.78ms            & 40.8M \\
            SG with DI  & \textbf{72.84}  & 180   & 2.42ms            & 40.8M \\
        \end{tabular}
         %\vspace{3mm}
        \caption{\footnotesize Results on CIFAR-100}
        \label{tab:cifar_accu}
	%\vspace{-10pt}
\end{table}
\begin{table}[t]
	    \centering\footnotesize
        \begin{tabular}{l | l | l | l }
            Mismatch       & Correct SC & Correct FC & Correct Combined \\
            \hline
            537            & 189        & 132        & 113 \\
        \end{tabular}
         %\vspace{3mm}
        \caption{\footnotesize Error analysis on SGNet model at epoch 180 }
        \label{tab:cifar_ep_180}
	\vspace{-5pt}
\end{table}

From Table~\ref{tab:cifar_accu}, we can see that SGNet with DI achieves the best performance, improving the original VGG-16 by $0.69\%$. SGNet with TSI also performs better than the original network, closely trailing behind DI by $0.06\%$. In fact, from all test results at the end of each epoch, DI consistently outperforms TSI. To analyze why DI performs better, we take the model from epoch $180$. As introduced in Section~\ref{sect:method}, both inference strategies share the same model with the only difference being the inference algorithm. 

In Table~\ref{tab:cifar_ep_180}, we show the error analysis. The SCB and FCB in the model generate super-class prediction and finer class prediction respectively. The mismatch in Table~\ref{tab:cifar_ep_180} is caused by the predicted finer class not being not under the predicted super-class. In all $10,000$ testing images, $537$ samples of such mismatch are found. Of the $537$ mismatched predictions, FCB is proven to be correct in $132$ of those images, but SCB has a better result with $189$ correct predictions. In TSI, only $113$ images are predicted correctly after combining both SCB and FCB results, which is less than DI. Therefore, if the finer class accuracy is the only thing that matters, then DI performs better. Note that it has more super-class prediction error. TSI actually makes less serious mistakes in prediction. So it really comes down to the application that determines which inference is better. If the consequence of a super-class error is severe, then TSI is actually preferred. 

\begin{figure}[t]
	\includegraphics[width=0.8\columnwidth]{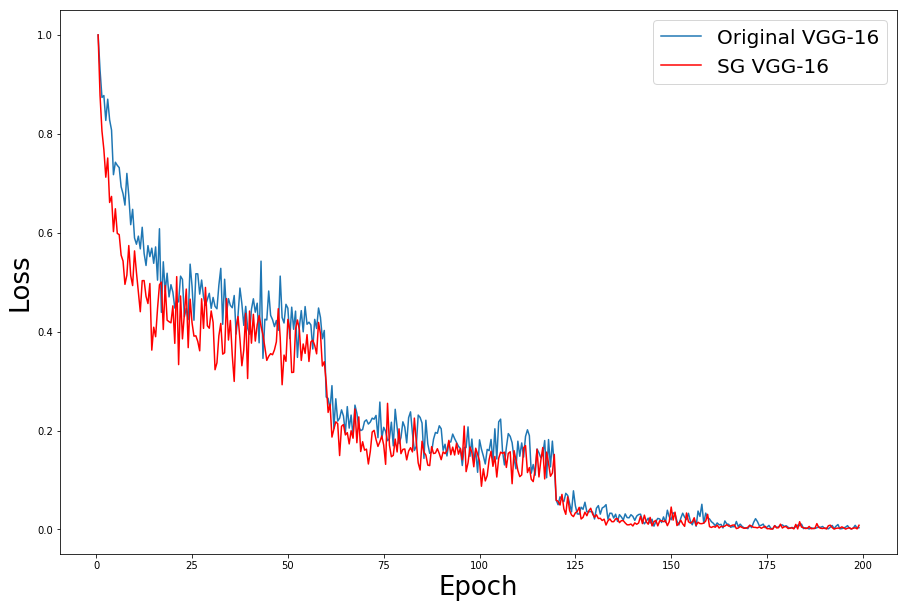}
	\centering
	\caption{Training loss on CIFAR-100}
	\label{fig:cifar-loss}
	\vspace{-5pt}
\end{figure}

In Table~\ref{tab:cifar_accu}, we also show the number of parameters in the networks we experimented with. The SGNet slightly increases the model size from $34.0$ million parameters to $40.8$ million. Correspondingly, the inference time is also increased a little bit.
%SGNet with DI increases the inference time by $17.5\%$ at $2.42$ms. Due to more complex inference algorithm, SGNet with TSI has a inference time of $2.78$ms. 
From the test result, we can see that SGNet achieves better performance than the original network. The loss curve is shown in Figure~\ref{fig:cifar-loss}. The loss values are normalized to $[0, 1]$ for the convenience of comparison. Figure~\ref{fig:cifar-loss} indicates that our model converges faster than the original VGG-16, especially at the early stage of training.

\subsection{Detection Datasets}
\noindent
\textbf{MS COCO} \cite{lin2014microsoft} is a large scale image dataset that can be used for object detection, object segmentation, etc. For object detection task, it has 80 object categories. It can be divided into 12 super classes, as shown in Table~\ref{tab:coco_sc_fc}. We can see that each super-class has a different number of finer classes. We use MS COCO's {\fontfamily{qcr}\selectfont 2014 train + 2014 val - 2014 minival} as the training set and {\fontfamily{qcr}\selectfont 2014 minival} as the validation set. The performance is evaluated using MS COCO's metrics APs at different IoUs (from $0.50$ to $0.95$) and APs for different object sizes.

\begin{table*}[t]
	    \centering\footnotesize
        \begin{tabular}{l|c}
            Super Class & Finer Class \\ \hline
            person     & person  \\ 
            vehicle    & bicycle, car, motorcycle, airplane, bus, train, truck, boat \\
            outdoor    & traffic light, fire hydrant, stop sign, parking meter, bench \\
            animal     & bird, cat, dog, horse, sheep, cow, elephant, bear, zebra, giraffe \\
            accessory  & backpack, umbrella, handbag, tie, suitcase \\
            sports     & frisbee, skis, snowboard, sports ball, kite, baseball bat, baseball glove, skateboard, surfboard, tennis racket \\
            kitchen    & bottle, wine glass, cup, fork, knife, spoon, bowl \\
            food       & banana, apple, sandwich, orange, broccoli, carrot, hot dog, pizza, donut, cake \\
            furniture  & chair, couch, potted plant, bed, dining table, toilet \\
            electronic & tv, laptop, mouse, remote, keyboard, cell phone \\
            appliance  & microwave, oven, toaster, sink, refrigerator \\
            indoor     & book, clock, vase, scissors, teddy bear, hair drier, toothbrush
        \end{tabular}
         \vspace{3mm}
        \caption{\footnotesize Super-class and finer class groups for MS COCO}
        \label{tab:coco_sc_fc}
	\vspace{-10pt}
\end{table*}

\subsection{Detection Model Architecture}
We implement our network based on Faster RCNN \cite{ren2015faster}. For the backbone network, we chose VGG-16 \cite{simonyan2014very}. We increased the output channel of the classification branch in Faster RCNN. As a two-step detector, Faster RCNN generates $C$ scores for each RoI as shown in Figure~\ref{plots:ts_det}, where $C$ is the number of classes. Therefore, in the original model, $C$ is $81$, background class plus $80$ classes. In our method, we extend $C$ to $94$ ($=13+81$), where $13$ is one \textit{background} class plus $12$ super-classes. Then, we split the final $94$-element array into super-class and finer class portions. 

In this detection task, we adopt a simpler learning rate strategy. It starts with $10^{-2}$ and decays every $5$ epochs with the decay rate of $0.1$. When reading the image, we use a scale of $800$ and the maximum size is $1200$ pixels.

\subsection{Detection Results}
The detection results on MS COCO are reported in Table~\ref{tab:coco_ap}. SGNet achieves better overall AP by $0.6\%$ compared to the original VGG-16. It is evident that our model has greater improvements when the IoU threshold is smaller. At an IoU of $0.50$, our model improves the AP by $1.1\%$. In addition, SGNet has the best effect when detecting large objects. This is caused by the components of the final loss in the training phase. In SGNet, the loss is the sum of L\textsubscript{sc}, L\textsubscript{fc}, and L\textsubscript{bbox}. Two terms in the loss are focused on training classification tasks. Therefore, the weight on the bounding box training is subsequently reduced. As shown in Table~\ref{tab:coco_ap}, the size and the inference time of the proposed detection model are almost the same as the original VGG-16 model, while the average precision is increased by $0.6\%$.

\begin{table}[t]
	    \centering\footnotesize
        \begin{tabular}{l | r r}
            Model                  &     VGG-16     & SGNet   \\
            \hline
            Detect Time            & 43.0ms         & 43.4ms  \\
            \# Params              & 138.06M        & 138.11M \\
            \hline
            AP                     & 28.6           & 29.2   \\ 
            AP\textsubscript{50}   & 48.6           & 49.7   \\
            AP\textsubscript{75}   & 30.3           & 30.8   \\
            \hline
            AP\textsubscript{S}    & 12.8           & 12.5   \\ 
            AP\textsubscript{M}    & 32.5           & 33.2   \\
            AP\textsubscript{L}    & 40.3           & 41.2   \\
        \end{tabular}
         \vspace{3mm}
        \caption{\footnotesize Results on MS COCO. VGG-16 and SGNet achieve best performance at epoch 15 and epoch 25 respectively. All experiments are carried out on one NVIDIA TITAN Xp GPU.}
        \label{tab:coco_ap}
	\vspace{-10pt}
\end{table}

%We increased the performance with little cost in model size and inference time. 
%As shown in Table~\ref{tab:coco_ap}, the size and the inference time of the proposed detection model are almost the same as the original VGG-16 model, while the average precision is increased by $0.6\%$. When compared to the inference time, which only increased from $43.0$ ms to $43.4$ ms, resulting in $23$ FPS, we can conclude an improvement in the overall performance.

\section{Conclusion}
In this paper, we have proposed a super-class guided network to integrate high-level knowledge into current image classification and object detection models. The proposed model integrates feature maps from both SCB and FCB to guide the finer class prediction. By creating an SCB model, we are able to add high-level knowledge hierarchy in the form of two-level class annotations. Extensive experiments have been performed on CIFAR-100 and MS COCO dataset. The results demonstrate the proposed SGNet can improve the performance of both image classification and object detection. The proposed network can be easily plugged into most existing models.

\section*{Acknowledgement}
The work was supported in part by The National Aeronautics and Space Administration (NASA) under grant no. 80NSSC20M0160, and the Natural Sciences and Engineering Research Council of Canada (NSERC) under grant RGPIN-2021-04244.
\balance 

\balance 

% that's all folks
\end{document}